\documentclass[letterpaper]{article}
\usepackage{authblk}
\usepackage{amsfonts}
\usepackage{amsmath}
\usepackage{amsthm}
\usepackage{amssymb}
\usepackage{bbold}
\usepackage{blochsphere}
\usepackage{booktabs}
\usepackage{circuitikz}
\usepackage{extdash}
\usepackage{multirow}
\usepackage{pgfplots}
\usepackage{physics}
\usepackage{quantikz}
\usepackage{siunitx}
\usepackage{standalone}
\usepackage{subcaption}
\usepackage{tablefootnote}
\usepackage{tikz}
\usepackage{times}
\usepackage{xspace}
\usepackage{mathtools}
\usepackage{nicefrac}
\usepackage{fullpage}
\usepackage[ruled]{algorithm2e}

\makeatletter
\renewcommand*\env@matrix[1][c]{\hskip -\arraycolsep
  \let\@ifnextchar\new@ifnextchar
  \array{*\c@MaxMatrixCols #1}}
\makeatother

\title{A Quantum Algorithm for Computing All Diagnoses of a Switching Circuit}

\author{Alexander Feldman}
\author{Johan de Kleer}
\author{Ion Matei}
\affil{Xerox PARC, Intelligent Systems Lab\\
3333 Coyote Hill Road, Palo Alto, CA 94304\\
\texttt{\small\string{afeldman,dekleer,imatei\string}@parc.com}}
\date{}

\usetikzlibrary{shapes}

\makeatletter
\definecolor{antiquewhite}{rgb}{0.98, 0.92, 0.84}
\definecolor{champagne}{rgb}{0.97, 0.91, 0.81}
\definecolor{darkkhaki}{rgb}{0.74, 0.72, 0.42}
\definecolor{pastelgray}{rgb}{0.81, 0.81, 0.77}
\definecolor{whitesmoke}{rgb}{0.96, 0.96, 0.96}
\definecolor{tearose(rose)}{rgb}{0.96, 0.76, 0.76}

\input{figures/pgflibrarystd_gates.code}
\makeatother

\theoremstyle{definition}

\theoremstyle{plain}

\newcommand{\cadical}{\textsc{CaDiCaL}\xspace}
\newcommand{\qiskit}{\textsc{Qiskit}\xspace}

\definecolor{antiquewhite}{rgb}{0.98, 0.92, 0.84}
\definecolor{pastelgray}{rgb}{0.81, 0.81, 0.77}
\definecolor{champagne}{rgb}{0.97, 0.91, 0.81}
\definecolor{cream}{rgb}{1.0, 0.99, 0.82}
\definecolor{darkkhaki}{rgb}{0.74, 0.72, 0.42}
\definecolor{tearose(rose)}{rgb}{0.96, 0.76, 0.76}

\newcommand{\pr}[1]{\ensuremath{\mbox{Pr}(#1)}}

\begin{document}

\maketitle

\begin{abstract}
  Faults are stochastic by nature while most man-made systems, and
  especially computers, work deterministically. This necessitates the
  linking of probability theory with mathematical logics, automata,
  and switching circuit theory. This paper provides such a connecting
  via quantum information theory which is an intuitive approach as
  quantum physics obeys probability laws.

  In this paper we provide a novel approach for computing diagnosis
  of switching circuits with gate-based quantum computers. The
  approach is based on the idea of putting the qubits representing
  faults in superposition and compute all, often exponentially many,
  diagnoses simultaneously.

  We empirically compare the quantum algorithm for diagnostics to an
  approach based on SAT and model-counting. For a benchmark of
  combinational circuits we establish an error of less than one
  percent in estimating the true probability of faults.
\end{abstract}

\section{Introduction}

Diagnostics of complex systems with many components is a challenging
and important topic \cite{dekleer1987diagnosing}. Problems in
Model-Based Diagnosis (MBD) are often NP-hard or worse
\cite{eiter1995complexity}. Quantum computing is a modern and
promising approach to solving a range of algorithmic challenges
\cite{montanaro2016quantum}. Researchers have already studied the problem of
finding a single minimal\-/cardinality diagnosis with an adiabatic
quantum computer \cite{ortiz19readiness}. In this paper we propose a
novel algorithm for the simultaneous computation of all diagnoses of a
system by using a traditional gate-based quantum computer.

The diagnostic problem we solve in this paper is framed as computing a
probability distribution function for each fault that can occur in a
switching circuit. This is driven by the fact that both in physics and
in complex systems, faults are stochastic. Classical computers and
switching circuits are deterministic. The result of
a classical quantum computation is always a probability distribution
function. Therefore, the mathematical framework defined by a quantum
computer can be used for combining switching circuits and probability.

The diagnostic problem we solve computes a probability distribution
function of each fault. The complexity of this problem is, to the best
of our knowledge, unknown. We hypothesize that the problem is
\#P-hard, the same as counting all solutions of a Disjunctive Normal
Form (DNF) formula. To analyze the correctness of our quantum
algorithm we compare it to a classical one based on enumerating all
solutions via satisfiability \cite{biere2009handbook} and blocking
clauses.

We have empirically evaluated the efficiency of the quantum diagnostic
algorithm on a benchmark of circuit families. Even in simulation, the
algorithm can handle surprisingly large circuits. We have had success
with circuits consisting of up to 30 gates. All experiments resulted
in a small error, typically within less than $1\%$ compared to the
golden standard computed by deterministic model counting.

The algorithm we present does not add to complexity theory. Although,
theoretically, one can use a quantum algorithm for model-counting and
3-SAT, in practice, the proposed algorithm is only for
approximation. The approximation error is arbitrarily large,
especially, when the underling SAT problem is in the phase-transition
region \cite{gent1994sat}. We discuss this in more detail in
Sec.~\ref{sec:discussion}.

\section{Preliminaries}

This text explains letters and symbols and helps with the intuitive
understanding of Boolean circuits. For precise definitions of Boolean
circuits, see the mathematical introduction to circuit complexity by
H. Vollmer~\cite{vollmer2013introduction}. For a thorough treaty
on the subject of quantum circuits, the reader is referred to the book
on quantum computing by M. Nielsen and I. Chuang
\cite{nielsen2000quantum}.

\subsection{Switching Circuits}

A switching circuit $C$ implements a multi-output Boolean function
$f_C: \{0, 1\}^m \to \{0, 1\}^n$. The $m$ arguments of $f_C$ are
primary inputs in $C$ and are denoted as
$X = \{x_1, x_2, \ldots, x_n\}$. Similarly, the $n$ results of $f_C$
are primary outputs in $C$ and are specified as
$Y = \{y_1, y_2, \ldots, y_m\}$.

\begin{figure}[hbt]
  \centering
  \begin{circuitikz}
  \draw (0, 0) node {NOT};
  \draw (2.5, 0) node {AND};
  \draw (5, 0) node {OR};
  \draw (7.5, 0) node {XOR};

  \draw (0, -1) node[not gate, fill=antiquewhite] (not) {};
  \draw (2.5, -1) node[and gate, fill=antiquewhite] (and) {};
  \draw (5, -1) node[or gate, fill=antiquewhite] (or) {};
  \draw (7.6, -1) node[xor gate, fill=antiquewhite] (xor) {};

  \draw (0, -2) node {$o \leftrightarrow \neg{i}$};
  \draw (2.5, -2) node {$o \leftrightarrow i_1 \wedge i_2$};
  \draw (5, -2) node {$o \leftrightarrow i_1 \vee i_2$};
  \draw (7.6, -2) node {$o \leftrightarrow i_1 \oplus i_2$};

  \draw (not.out) -- ++(0.25, 0) node[anchor=south east, inner xsep=0] {$o$};
  \draw (not.in) -- ++(-0.25, 0) node[anchor=south west, inner xsep=0] {$i$};

  \draw (and.out) -- ++(0.25, 0) node[anchor=south east, inner xsep=0] {$o$};
  \draw (and.in 1) -- ++(-0.25, 0) node[anchor=south west, inner xsep=0] {$i_1$};
  \draw (and.in 2) -- ++(-0.25, 0) node[anchor=south west, inner xsep=0] {$i_2$};

  \draw (or.out) -- ++(0.25, 0) node[anchor=south east, inner xsep=0] {$o$};
  \draw (or.in 1) -- ++(-0.25, 0) node[anchor=south west, inner xsep=0] {$i_1$};
  \draw (or.in 2) -- ++(-0.25, 0) node[anchor=south west, inner xsep=0] {$i_2$};

  \draw (xor.out) -- ++(0.35, 0) node[anchor=south east, inner xsep=0] {$o$};
  \draw (xor.in 1) -- ++(-0.35, 0) node[anchor=south west, inner xsep=0] {$i_1$};
  \draw (xor.in 2) -- ++(-0.35, 0) node[anchor=south west, inner xsep=0] {$i_2$};
\end{circuitikz}%
  \caption{The standard basis}
  \label{fig:standard_basis}
\end{figure}
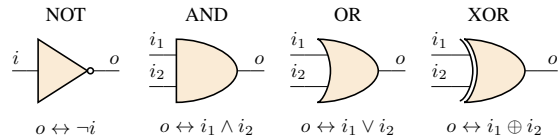

A switching circuit is a graphical network that connects gates with
wires. Each gate implement some (typically small) Boolean
function. This set of elementary Boolean functions is called the basis
of the circuit. Figure~\ref{fig:standard_basis} shows a common set of
gates and their corresponding Boolean formulas that are used often in
computer science and VLSI design. It contains of two-input AND, OR,
and XOR gates and an inverter. With this minimalistic basis,
multi-input AND and OR gates are implemented by chaining; and NAND,
and NOR gates are implemented by appending invertors to AND and OR
gates, respectively.

Some common Boolean functions are negation ($\neg$), disjunction
($\vee$), conjunction ($\wedge$), exclusive or ($\oplus$),
implication\footnote{This paper, similar to many others, shares the
  same symbol ($\rightarrow$) for implication and for function
  mapping. The use is clear from the context.} ($\rightarrow$), and
equivalence ($\leftrightarrow$). This paper uses everywhere infix, as
opposed to prefix, notation. For example, $p \vee q$ is used instead
of $\vee\left(p, q\right)$.

We also use equivalence ($\leftrightarrow$) instead of the equal sign
($=$) to specify Boolean functions. The function output is on the left
while the inputs are on the right. For example, the Boolean function
$r = p \vee q$ is written as $r \leftrightarrow p \vee q$. When there
are multiple outputs, we give a formula for each one of them.

\begin{figure}[hbt]
  \centering
  \begin{circuitikz}
  \draw[rounded corners, dashed, fill=whitesmoke] (-2.375, 0.75) rectangle +(2.5, -3.5);
  \draw[rounded corners, dashed, fill=whitesmoke] (0.375, -0.5) rectangle +(2.5, -3.5);

  \node[anchor=south west, inner ysep=4pt] at (-2.375, -2.75) {Half-Adder};
  \node[anchor=north west, inner ysep=4pt] at (0.375, -0.5) {Half-Adder};

  \draw (0, 0) node[and gate, anchor=out, fill=pastelgray] (a1) {};
  \draw (0, -1.5) node[xor gate, anchor=out, fill=pastelgray] (x1) {};

  \draw (1.175, -1.5) node[and gate, anchor=in 1, fill=champagne] (a2) {};
  \draw (1.175, -3) node[xor gate, anchor=in 1, fill=champagne] (x2) {};

  \draw (4.175, 0) node[or gate, anchor=in 1, fill=darkkhaki] (o) {};

  \draw (a1.in 1) -- +(-0.5, 0) node[draw, fill=black, circle, inner sep=1.5pt] (i1) {} --
                     +(-1.25, 0) node[inner xsep=0, inner ysep=2pt, align=left, anchor=south west] {$i_1$};
  \draw (x1.in 2) -- +(-0.25, 0) node[draw, fill=black, circle, inner sep=1.5pt] (i2) {} --
                     +(-1.25, 0) node[inner xsep=0, inner ysep=2pt, align=left, anchor=south west] {$i_2$};
  \draw (x2.in 2) -- +(-0.25, 0) node[draw, fill=black, circle, inner sep=1.5pt] (ci) {} --
                     +(-1.5, 0) node[inner xsep=0, inner ysep=2pt, align=left, anchor=south west] {$c_i$};

  \draw (a1.out) -- (o.in 1) node[inner xsep=0, inner ysep=2pt, anchor=south, midway] {$z_1$};
  \draw (o.out) -- +(0.5, 0) node[inner xsep=0, inner ysep=2pt, align=right, anchor=south east] {$c_o$};
  \draw (x2.out) -- +(0.625, 0) node[inner xsep=0, inner ysep=2pt, align=right, anchor=south east] {$\sigma$};

  \draw (a2.in 1) -- +(-0.5, 0) node[draw, fill=black, circle, inner sep=1.5pt] (z2) {} -- (x1.out);
  \draw (a2.out) -| ($(a2.out)!0.5!(o.in 1)$) node[inner sep=2pt, pos=0.8, anchor=west] {$z_3$} |- (o.in 2);
                     
  \draw (a1.in 2) -| (i2.center);
  \draw (a2.in 2) -| (ci.center);
  \draw (x1.in 1) -| (i1.center);
  \draw (x2.in 1) -| (z2.center);

  \draw node[inner xsep=0pt, inner ysep=4pt, anchor=south] at (z2) {$z_2$};
\end{circuitikz}
  \caption{A full-adder}
  \label{fig:full_adder}
\end{figure}
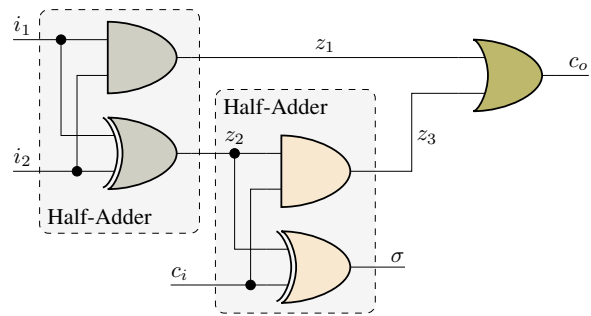

Figure~\ref{fig:full_adder} shows a simple and frequently used
switching circuit that is used for adding the two binary numbers $i_1$
and $i_2$ and a carry input bit $c_i$. The output is found in the sum
bit $\sigma$ and in the carry output $c_o$. Notice that there are two
identical subcircuits in Figure~\ref{fig:full_adder}. These are the
two half-adders.

An assignment to the primary inputs or outputs of a switching circuit
is a set of variable/value pairs. The values are the Boolean constants
$0$ or $1$. An assignment can be written as a propositional
conjunction. For example assigning $i_1 = 1$, $i_2 = 0$, and $c_i = 1$
is written as $\neg{i_1} \wedge \neg{i_2} \wedge c_i$.

\subsection{Quantum Circuits}

Although, on the surface, quantum circuits can be reminiscent of the
switching ones described above, there are multiple important
differences. The biggest one is, that while the state of a Boolean
circuit is a Boolean vector over all wires, the state of a quantum
circuit is a superposition over all quantum bits. The quantum state is
of size, exponential to the number of quantum bits.

A quantum circuit is a set of quantum wires
$\psi_1, \psi_2, \ldots, \psi_n$ that go through quantum gates. In
reality a quantum wire does not have to look anything like a piece of
copper metal but can be, for example laser light, or even some passage
of time.

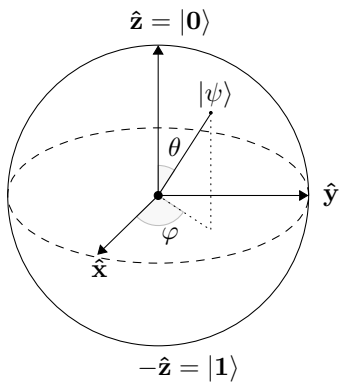
\begin{figure}[hbt]
  \centering
  \begin{tikzpicture}[line cap=round, line join=round, >=Triangle]
  \clip(-2.19,-2.49) rectangle (2.66,2.58);
  \draw [shift={(0,0)}, lightgray, fill, fill opacity=0.1] (0,0) -- (56.7:0.4) arc (56.7:90.:0.4) -- cycle;
  \draw [shift={(0,0)}, lightgray, fill, fill opacity=0.1] (0,0) -- (-135.7:0.4) arc (-135.7:-33.2:0.4) -- cycle;
  \draw(0,0) circle (2cm);
  \draw [rotate around={0.:(0.,0.)},dash pattern=on 3pt off 3pt] (0,0) ellipse (2cm and 0.9cm);
  \draw (0,0) -- (0.7,1.09);
  \draw [->] (0,0) -- (0,2);
  \draw [->] (0,0) -- (-0.81,-0.79);
  \draw [->] (0,0) -- (2,0);
  \draw [dotted] (0.7,1)-- (0.7,-0.46);
  \draw [dotted] (0,0)-- (0.7,-0.46);
  \draw (-0.08,-0.3) node[anchor=north west] {$\varphi$};
  \draw (0.01,0.9) node[anchor=north west] {$\theta$};
  \draw (-1.01,-0.72) node[anchor=north west] {$\mathbf {\hat{x}}$};
  \draw (2.07,0.3) node[anchor=north west] {$\mathbf {\hat{y}}$};
  \draw (-0.5,2.6) node[anchor=north west] {$\mathbf {\hat{z}=|0\rangle}$};
  \draw (-0.4,-2) node[anchor=north west] {$-\mathbf {\hat{z}=|1\rangle}$};
  \draw (0.4,1.65) node[anchor=north west] {$|\psi\rangle$};
  \scriptsize
  \draw [fill] (0,0) circle (1.5pt);
  \draw [fill] (0.7,1.1) circle (0.5pt);
\end{tikzpicture}
  \caption{Bloch sphere}
  \label{fig:blochsphere}
\end{figure}

The simplest quantum circuit consists of one quantum wire $\psi_1$ and
one or more quantum gates that act on this single wire. The state of
the quantum circuit can be illustrated as a position on a fictional
Bloch sphere (see Figure~\ref{fig:blochsphere}). Using P. Dirac's
notation, the state of the circuit is
$\ket{\psi} = \alpha \ket{0} + \beta \ket{1}$, where
$\alpha, \beta \in \mathbb{C}$. In this bootstrapping example all kets
are vectors of size two, $\ket{0}$ is
$\begin{bsmallmatrix}1\\0\end{bsmallmatrix}$, and $\ket{1}$ is
$\begin{bsmallmatrix}0\\1\end{bsmallmatrix}$.

The state of a quantum circuit with two qubits is a superposition with
four terms: $\ket{\psi_1 \psi_2} = \alpha_{00}\ket{00} + \alpha_{01}\ket{01} +
\alpha_{10}\ket{10} + \alpha_{11}\ket{11}$ where
$\alpha_{00}, \alpha_{01}, \alpha_{10}, \alpha_{11} \in \mathbb{C}$.

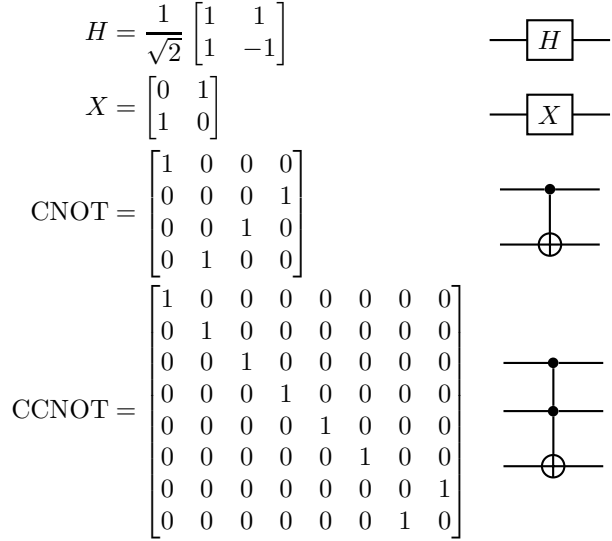
\begin{figure}[hbt]
  \centering

\newsavebox{\gateh}%
\savebox{\gateh}%
{%
\begin{quantikz}%
\qw & \gate{H} & \qw%
\end{quantikz}%
}%
\newsavebox{\gatex}%
\savebox{\gatex}%
{%
\begin{quantikz}%
\qw & \gate{X} & \qw%
\end{quantikz}%
}%
\newsavebox{\gatecnot}%
\savebox{\gatecnot}%
{%
\begin{quantikz}%
\qw & \ctrl{1} & \qw \\%
\qw & \targ{}  & \qw%
\end{quantikz}%
}%
\newsavebox{\gateccnot}%
\savebox{\gateccnot}%
{
\begin{quantikz}%
\qw & \ctrl{1} & \qw \\%
\qw & \ctrl{1} & \qw \\%
\qw & \targ{}  & \qw%
\end{quantikz}%
}%
\renewcommand{\tabcolsep}{1pt}
\begin{tabular}{r@{$\ =\ $}lc}%
$H$%
&%
$\displaystyle%
\frac{1}{\sqrt{2}}%
\begin{bmatrix}%
1 & 1 \\%
1 & -1 %
\end{bmatrix}%
$%
\vspace{1mm}%
&%
\usebox\gateh%
\\%
$X$%
&%
$\displaystyle%
\begin{bmatrix}%
0 & 1 \\%
1 & 0%
\end{bmatrix}%
$%
\vspace{1mm}%
&%
\usebox\gatex%
\\
$\mathrm{CNOT}$
&%
$\displaystyle
\begin{bmatrix}
1 & 0 & 0 & 0 \\
0 & 0 & 0 & 1 \\
0 & 0 & 1 & 0 \\
0 & 1 & 0 & 0 \\
\end{bmatrix}
$%
\vspace{1mm}%
&%
\usebox\gatecnot%
\\%
$\mathrm{CCNOT}$
&%
$\displaystyle
\begin{bmatrix}
1 & 0 & 0 & 0 & 0 & 0 & 0 & 0 \\
0 & 1 & 0 & 0 & 0 & 0 & 0 & 0 \\
0 & 0 & 1 & 0 & 0 & 0 & 0 & 0 \\
0 & 0 & 0 & 1 & 0 & 0 & 0 & 0 \\
0 & 0 & 0 & 0 & 1 & 0 & 0 & 0 \\
0 & 0 & 0 & 0 & 0 & 1 & 0 & 0 \\
0 & 0 & 0 & 0 & 0 & 0 & 0 & 1 \\
0 & 0 & 0 & 0 & 0 & 0 & 1 & 0 \\
\end{bmatrix}
$%
\vspace{1mm}%
&%
\usebox\gateccnot%
\end{tabular}%
\renewcommand{\tabcolsep}{6pt}

  \caption{Quantum gates and their corresponding matrices}
  \label{fig:quantum_basis}
\end{figure}

Three types of quantum gates are sufficient for illustrating the
quantum algorithms of this paper: (1) Pauli-X, (2) Hadamard,
and (3) CNOT gate. The symbols and matrices of these gates are shown in
Figure~\ref{fig:quantum_basis}.

\section{Algorithms}

After a brief introduction to fault-modeling and conditional fault
probability, we present a SAT-based algorithm for diagnostic
model-counting and then we proceed with the main contribution of this
paper: the quantum algorithm for circuit diagnostics.

\subsection{Fault-Modeling}

Our approach to diagnosis is to automatically modify the input circuit
$C$ and to use a generic reasoning algorithm which does not use the
notion of faults. The method we employ is based on sub-circuit
rewriting. Depending on the exact type of rewriting, it is possible to
model stuck-at faults, gates behaving like other gates (for example an
AND-gate, behaving like an OR-gate), short-circuits between two or
more wires, and others.

In this paper, similar to Automatic Test-Pattern Generation (ATPG)
\cite{feldman21atpg}, we are interested in stuck-at-faults. Unlike in
ATPG where the fault location can be both at the stem and at the
branches of a fan-out we place stuck-at faults only at gate
outputs. To model for a stuck-at-one gate, it is enough to insert an
OR-gate at the fault location. The second input of the OR-gate becomes
a special type of a primary input, called \textit{assumable}.

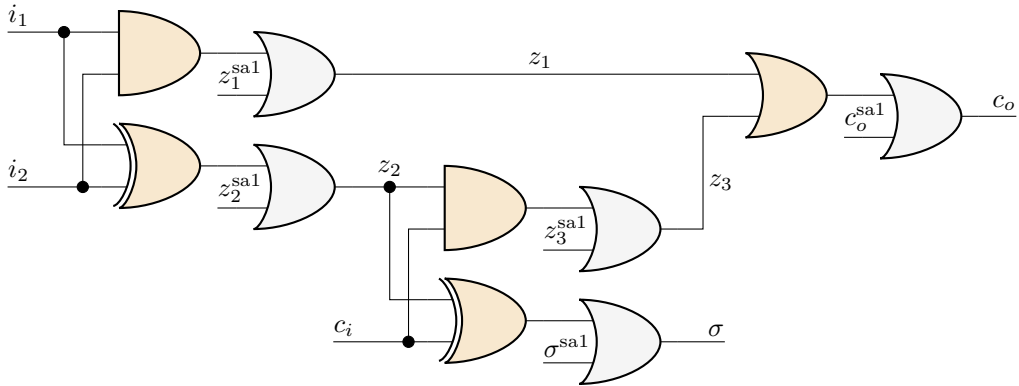
\begin{figure*}[hbt]
  \centering
  \begin{circuitikz}
  \draw (0, 0) node[and gate, anchor=out, fill=champagne] (a1) {};
  \draw (0, -1.5) node[xor gate, anchor=out, fill=champagne] (x1) {};
 
  \draw (a1.in 1) -- +(-0.5, 0) node[draw, fill=black, circle, inner sep=1.5pt] (i1) {} --
                     +(-1.25, 0) node[inner xsep=0, inner ysep=2pt, align=left, anchor=south west] {$i_1$};
  \draw (x1.in 2) -- +(-0.25, 0) node[draw, fill=black, circle, inner sep=1.5pt] (i2) {} --
                     +(-1.25, 0) node[inner xsep=0, inner ysep=2pt, align=left, anchor=south west] {$i_2$};
  \draw (a1.in 2) -| (i2.center);
  \draw (x1.in 1) -| (i1.center);

  \draw (a1.out) -- +(0.25, 0) node[or gate, anchor=in 1, fill=whitesmoke] (a1 sa1) {};
  \draw (x1.out) -- +(0.25, 0) node[or gate, anchor=in 1, fill=whitesmoke] (x1 sa1) {};

  \draw (x1 sa1.out) -- +(1, 0) node[and gate, anchor=in 1, fill=champagne] (a2) {};
  \draw ($(a2.out) + (0, -1.5)$) node[xor gate, anchor=out, fill=champagne] (x2) {};

  \draw ($(a2.in 1) + (-0.5, 0)$) node[draw, fill=black, circle, inner sep=1.5pt] (z2) {};
  \draw (x2.in 2) -- +(-0.25, 0) node[draw, fill=black, circle, inner sep=1.5pt] (ci) {} --
                     +(-1.25, 0) node[inner xsep=0, inner ysep=2pt, align=left, anchor=south west] {$c_i$};

  \draw (a2.in 2) -| (ci.center);
  \draw (x2.in 1) -| (z2.center);

  \draw (a2.out) -- +(0.25, 0) node[or gate, anchor=in 1, fill=whitesmoke] (a2 sa1) {};
  \draw (x2.out) -- +(0.25, 0) node[or gate, anchor=in 1, fill=whitesmoke] (x2 sa1) {};

  \draw (a1 sa1.out) -- ++(2.5, 0) node[inner xsep=0, inner ysep=2pt, anchor=south] {$z_1$} -- ++(2.5, 0) node[or gate, anchor=in 1, fill=champagne] (o) {};

  \draw (o.out) -- +(0.25, 0) node[or gate, anchor=in 1, fill=whitesmoke] (o sa1) {};

  \draw (o sa1.out) -- +(0.5, 0) node[inner xsep=0, inner ysep=2pt, align=right, anchor=south east] {$c_o$};
  \draw (x2 sa1.out) -- +(0.625, 0) node[inner xsep=0, inner ysep=2pt, align=right, anchor=south east] {$\sigma$};

  \draw (a2 sa1.out) -| ($(a2 sa1.out)!0.5!(o.in 1)$) node[inner sep=2pt, pos=0.8, anchor=west] {$z_3$} |- (o.in 2);

  \draw (a1 sa1.in 2) -- +(-0.25, 0) node[inner xsep=0, inner ysep=2pt, align=left, anchor=south west] {$z_1^\mathrm{sa1}$};
  \draw (x1 sa1.in 2) -- +(-0.25, 0) node[inner xsep=0, inner ysep=2pt, align=left, anchor=south west] {$z_2^\mathrm{sa1}$};
  \draw (a2 sa1.in 2) -- +(-0.25, 0) node[inner xsep=0, inner ysep=2pt, align=left, anchor=south west] {$z_3^\mathrm{sa1}$};
  \draw (x2 sa1.in 2) -- +(-0.25, 0) node[inner xsep=0, inner ysep=2pt, align=left, anchor=south west] {$\sigma^\mathrm{sa1}$};
  \draw (o sa1.in 2) -- +(-0.25, 0) node[inner xsep=0, inner ysep=2pt, align=left, anchor=south west] {$c_o^\mathrm{sa1}$};

  \draw node[inner xsep=0pt, inner ysep=4pt, anchor=south] at (z2) {$z_2$};
\end{circuitikz}
  \caption{A fault-augmented full-adder}
  \label{fig:full_adder_sa}
\end{figure*}

Figure~\ref{fig:full_adder_sa} shows the full-adder example from
Figure~\ref{fig:full_adder} with all outputs of the original five
gates allowed to be stuck-at-1. The five new assumable inputs are
$z_1^\mathrm{sa1}$, $z_2^\mathrm{sa1}$, $z_3^\mathrm{sa1}$,
$\sigma^\mathrm{sa1}$, and $c_o^\mathrm{sa1}$.

\subsection{Conditional Probability of Faults}

In the preceding section, we have shown, how computation of a
diagnosis can be reduced to finding compatible values for a special
subset of inputs, the fault inputs. In reality, however, faults show
stochastic behavior, so it is worth to establish a computational
framework for probabilities.

Connecting probabilities to switching circuits has been done by John
von Neumann \cite{vonneumann1956probabilistic}, while Parker and
McCluskey show how, from a circuit and the probability of each input
one can compute the probability of each output
\cite{parker1975probabilistic}. Parker and McCluskey show two
algorithms for computing the probabilities of the circuit outputs: one
that first computes all prime implicants of the Boolean function
modeling the switching circuit, and the other by starting from the
inputs and performing real-value arithmetics over the probability
values. It is of no surprise that the second algorithm has polynomial
complexity as forward propagation from inputs to outputs can be done
easily.

We extend the framework of K. Parker and E. McCluskey by allowing one
to compute the probability of any wire from any set of probabilities
we know. In particular, we are interested in computing the
probabilities of a subset of inputs, the fault inputs given the
primary inputs and the primary outputs of a circuit. For that we use
an NP-complete approach, i.e., we implicitly create a table with all
consistent values. Each row of the table is a diagnosis and each
column is a fault variable. To compute the probability of a fault we
divide the number of diagnoses in which the specified fault shows-up
by the total number of diagnoses.

Consider the circuit $C_f$ from Fig.~\ref{fig:full_adder_sa}, the
primary inputs assignment
$\alpha = \neg{i_1} \wedge \neg{i_2} \wedge c_i$, and a double-fault
injection $\gamma = \sigma^\mathrm{sa1} \wedge c_o^\mathrm{sa1}$. The
primary outputs are actually determined only by the fault $\gamma$ and
are $\beta = \sigma \wedge c_o$.
Table~\ref{tbl:full_adder_sa1_sat_diag} contains all diagnoses of
$C_f$, $\alpha$, and $\beta$; as computed by
Algorithm~\ref{alg:sat_diag}. There is a total of $22$ diagnoses and
their distribution resembles binomial as there is relatively little
masking in this example circuit.

\renewcommand{\tabcolsep}{3pt}
\begin{table}[htb]
  \begin{center}
    \begin{tabular}{cc}
      \begin{tabular}[t]{ccccc}
        \toprule
        $z_1^\mathrm{sa1}$ & $z_2^\mathrm{sa1}$ & $z_3^\mathrm{sa1}$ & $\sigma^\mathrm{sa1}$ & $c_o^\mathrm{sa1}$ \\
        \midrule
        $0$ & $0$ & $0$ & $0$ & $1$ \\
        $0$ & $0$ & $1$ & $0$ & $0$ \\
        $1$ & $0$ & $0$ & $0$ & $0$ \\
        $0$ & $0$ & $0$ & $1$ & $1$ \\
        $0$ & $0$ & $1$ & $0$ & $1$ \\
        $0$ & $0$ & $1$ & $1$ & $0$ \\
        $0$ & $1$ & $0$ & $1$ & $0$ \\
        $1$ & $0$ & $0$ & $0$ & $1$ \\
        $1$ & $0$ & $0$ & $1$ & $0$ \\
        $1$ & $0$ & $1$ & $0$ & $0$ \\
        $0$ & $0$ & $1$ & $1$ & $1$ \\
        $\vdots$ & $\vdots$ & $\vdots$ & $\vdots$ & $\vdots$ \\
        \midrule
      \end{tabular}
      &
      \begin{tabular}[t]{ccccc}
        \toprule
        $z_1^\mathrm{sa1}$ & $z_2^\mathrm{sa1}$ & $z_3^\mathrm{sa1}$ & $\sigma^\mathrm{sa1}$ & $c_o^\mathrm{sa1}$ \\
        \midrule
        $\vdots$ & $\vdots$ & $\vdots$ & $\vdots$ & $\vdots$ \\
        $0$ & $1$ & $0$ & $1$ & $1$ \\
        $0$ & $1$ & $1$ & $1$ & $0$ \\
        $1$ & $0$ & $0$ & $1$ & $1$ \\
        $1$ & $0$ & $1$ & $0$ & $1$ \\
        $1$ & $0$ & $1$ & $1$ & $0$ \\
        $1$ & $1$ & $0$ & $1$ & $0$ \\
        $0$ & $1$ & $1$ & $1$ & $1$ \\
        $1$ & $0$ & $1$ & $1$ & $1$ \\
        $1$ & $1$ & $0$ & $1$ & $1$ \\
        $1$ & $1$ & $1$ & $1$ & $0$ \\
        $1$ & $1$ & $1$ & $1$ & $1$ \\
        \midrule
        $\frac{12}{22}$ & $\frac{8}{22}$ & $\frac{12}{22}$ & $\frac{15}{22}$ & $\frac{12}{22}$ \\
        \bottomrule
      \end{tabular}
    \end{tabular}
  \end{center}
  \caption{Truth-table, diagnoses and conditional fault-probabilities
    for the full-adder running example and an
    observation\label{tbl:full_adder_sa1_sat_diag}}
\end{table}
\renewcommand{\tabcolsep}{6pt}

\subsection{An Algorithm for Computing Conditional Fault Probabilities Based on Satisfiability}

Algorithm~\ref{alg:sat_diag} is a classical reference algorithm for
calculating the conditional probability $P(f_i | C, \alpha)$ for each
$f_i \in F$. It converts the circuit $C$ to a Boolean formula in
Conjunctive Normal Form and repetitively calls a SAT solver to compute
satisfiable solutions that also happen to be diagnoses. After each
call to the SAT subroutine, a row is added to a table from which the
conditional probability can be computed and a blocking clause
is added to prevent the satisfiable solution $\omega$ from showing-up
again.
\begin{algorithm}[htb]
  \caption{\textsc{CircuitHealthSAT}}
  \label{alg:sat_diag}

  \SetKwInOut{Input}{Input}
  \SetKwInOut{Output}{Output}
  \SetKwInOut{Local}{Local Variables~}

  \Input{$C_f$, the fault-augmented input circuit with fault inputs $F = \{f_1, f_2, \ldots, f_n\}$}
  \Input{$\alpha$, primary inputs assignment}
  \Input{$\beta$, primary outputs assignment}
  \Output{$R$, conditional probabilities of faults}

  \BlankLine

  $B \gets \top$ \\
  \For{$f \in F$} {
    $n[f] \gets 0$ \\  
  }
  $d \gets 0$ \\
  \While{$\omega \gets \normalfont\textsc{SAT}(\normalfont\textsc{CNF}(C_f) \wedge \alpha \wedge \beta \wedge B)$} {
    \For{$f \in F$} {
      \If{$\normalfont\textsc{IsFault}(f)$} {
        $n[f] \gets n[f] + 1$ \\
      }
      $B \gets B \wedge \textsc{BlockingClause}(\omega)$ \\
      $d \gets d + 1$ \\
    }
  }
  $R \gets \left\{\frac{n[f]}{d} : f \in F\right\}$ \\
  \textbf{return} $R$
\end{algorithm}

The truth table does not need to be created explicitly and it is
sufficient to maintain a set of counters for each fault
variable. These counters are denoted as $n[f], f \in F$ where $F$ is
the set of all fault variables. The total number of satisfiable
assignments (diagnoses) is kept in $d$. To receive the probability of
a fault it is enought to divide the respective $n[f]$ by the total
number of diagnoses $d$.

\subsection{Quantum-Based Conditional Probability}

A Hilbert space representing a quantum system must satisfy the
normalization condition. The inner product of a vector with itself is
equal to one: $\bra{\psi}\ket{\psi} = 1$. The length of a vector in a
particular direction represents the ``probability amplitude'' of the
quantum system. The idea is to set all assumable inputs to a
superposition of $\ket{0}$ and $\ket{1}$ and to apply a quantum oracle
that is analogous to the Boolean circuit being diagnosed. The
collapsed probability amplitude readout is the a posteriori probability
distribution function of each fault given the observation $\alpha$ and
the circuit $\varphi$.

\subsubsection{Making the Quantum Oracle Circuit}

\begin{figure}[hbt]
  \centering
  \subcaptionbox{Inverter}[0.3\columnwidth]
  {%
    \begin{quantikz}
      \lstick{$\ket{\psi_1}$} & \ctrl{1} & \qw \\
      \lstick{$\ket{1}$}      & \targ{}  & \qw
    \end{quantikz}%
  }%
  \subcaptionbox{AND Gate}[0.3\columnwidth]
  {%
    \begin{quantikz}
      \lstick{$\ket{\psi_1}$} & \ctrl{2} & \qw \\
      \lstick{$\ket{\psi_2}$} & \ctrl{1} & \qw \\
      \lstick{$\ket{0}$}      & \targ{}  & \qw
    \end{quantikz}%
  }%
  \subcaptionbox{XOR Gate}[0.4\columnwidth]
  {%
    \begin{quantikz}
      \lstick{$\ket{\psi_1}$} & \ctrl{2} & \qw      & \qw \\
      \lstick{$\ket{\psi_2}$} & \qw      & \ctrl{1} & \qw \\
      \lstick{$\ket{0}$}      & \targ{}  & \targ{}  & \qw
    \end{quantikz}%
  }%
  \vspace{2.5mm}
  \subcaptionbox{OR Gate}[\columnwidth]
  {%
    \begin{quantikz}
      \lstick{$\ket{\psi_1}$} & \gate{X} & \ctrl{2} & \qw \\
      \lstick{$\ket{\psi_2}$} & \gate{X} & \ctrl{1} & \qw \\
      \lstick{$\ket{1}$}      & \qw      & \targ{}  & \qw
    \end{quantikz}%
  }%
  \caption{Quantum implementation of classical gates from the standard
  basis\label{fig:classical_quantum_implementations}}
\end{figure}
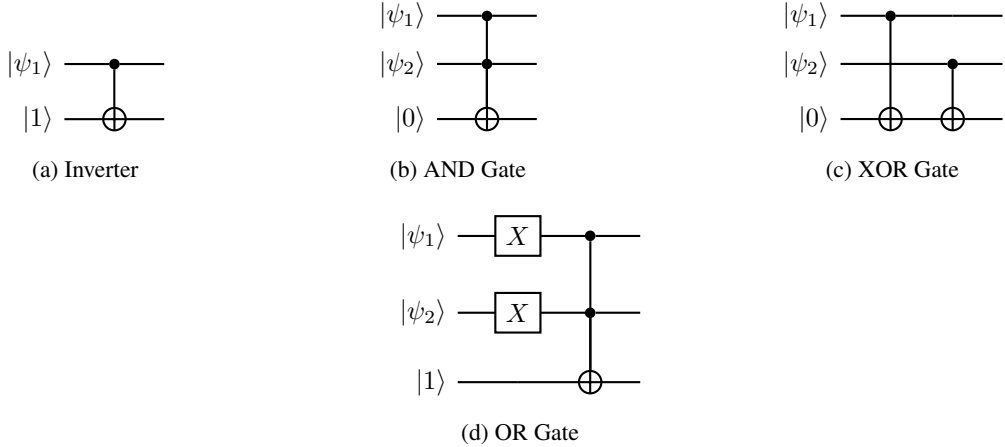

The first step is to create a quantum oracle by converting the
classical circuit to a quantum one. Algorithm~\ref{alg:make_oracle},
which is classical, shows this transformation. It starts by adding a
quantum wire for each primary input of the circuit. Then, for each
classical gate, a quantum sub-circuit is added to the quantum one. The
quantum circuits implementing the standard classical gates are shown
in Figure~\ref{fig:classical_quantum_implementations}.

\begin{algorithm}[htb]
  \caption{\textsc{MakeOracle}($C$)}
  \label{alg:make_oracle}

  \SetKwInOut{Input}{Input}
  \SetKwInOut{Output}{Output}
  \SetKwInOut{Local}{Local Variables~}

  \Input{$C$, the input circuit}
  \Output{$U$, the quantum circuit oracle}
  \Local{$M$, dictionary, Boolean variables to quantum wires map}

  \BlankLine
  $U \gets \emptyset$ \\
  $C = \textsc{SortGates}(C)$

  \For{$q \in \{1, 2, \ldots, m\}$} {
    \textsc{$M[q] \gets q$}\\
  }
  \For{$g \in \{1, 2, \ldots, n\}$} {
    \Switch{\textsc{GateType(g)}}
    {
      $\psi_1 = M[i_1], \psi_2 = M[i_2], \ldots$\\
      \Case{Inverter}
      {
        $U(i) \gets \mathrm{CCNOT}(I \otimes X)\ket{\psi_1 a_i}$\\
      }
      \Case{AND-gate}
      {
        $U(i) \gets \mathrm{CCNOT}\ket{\psi_1 \psi_2 a_i}$\\
      }
      \Case{OR-gate}
      {
        $U(i) \gets \mathrm{CCNOT}\ \mathrm{X}^3\ \ket{\psi_1 \psi_2 a_i}$\\
      }
      \Case{XOR-gate}
      {
        $U(i) \gets \mathrm{CNOT} (I \otimes \mathrm{CNOT})\ket{\psi_2 \psi_1 a_i}$\\
      }
    }
    $M[o] = a_i$\\
    $i \gets i + 1$\\
  }
  \Return{U}
\end{algorithm}

Algorithm~\ref{alg:make_oracle} first topologically sorts all gates in
the switching circuit. This can be done with a simple graph traversal
starting from the primary inputs and proceeding toward the primary
outputs. The map $M$ keeps the correspondences between the switching
circuit wires and the qubits in the quantum oracle. During the
initialization phase, Algorithm~\ref{alg:make_oracle} creates a qubit
for each primary input and for each fault input and places a
one-to-one entries for them in $M$.

During the main phase of Algorithm~\ref{alg:make_oracle}, each gate in
the input circuit $C$ is replaced with its corresponding quantum
sub-circuit (see
Figure~\ref{fig:classical_quantum_implementations}). The quantum
circuits are constructed such that the information on the input wires
is preserved, so it can be reused by downstream
gates. Algorithm~\ref{alg:make_oracle} also adds a new ancillary qubit
for each gate.

\begin{figure*}[hbt]
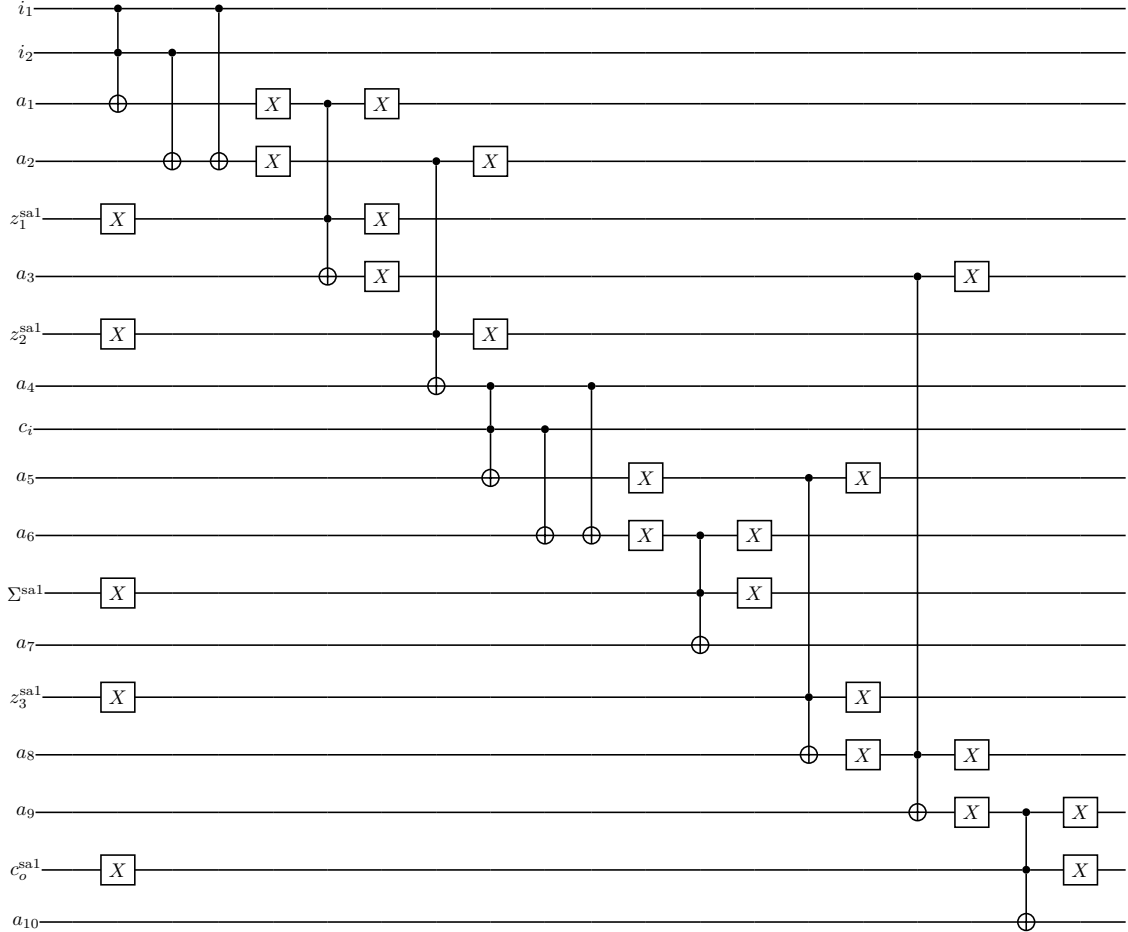

  \centering
  \includestandalone[scale=0.75]{figures/full_adder_sa1_oracle}%
  \caption{Quantum circuit oracle from the fault-augmented switching
    circuit shown in Figure~\ref{fig:full_adder_sa}}
  \label{fig:full_adder_sa_oracle}
\end{figure*}

Figure~\ref{fig:full_adder_sa_oracle} shows the result of applying
Algorithm~\ref{alg:make_oracle} on the full-adder running example from
Figure~\ref{fig:full_adder_sa}.

\subsubsection{Quantum Algorithm for Circuit Diagnosis}

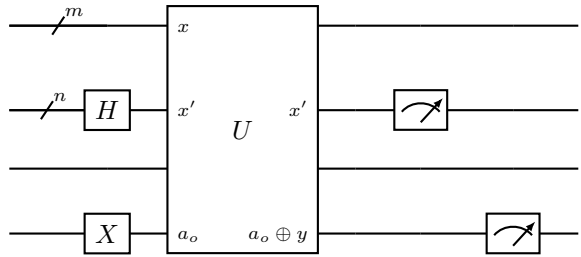
\begin{figure}[hbt]
\begin{center}
\begin{quantikz}
\qw &[5mm] \qw\qwbundle{m}      & \gate[wires=4][2cm]{U}\gateinput{$x$}           & \qw    & \qw      & \qw      & \qw \\
\qw &[5mm] \gate{H}\qwbundle{n} & \gateinput{$x^\prime$} \gateoutput{$x^\prime$}  & \qw    & \meter{} & \qw      & \qw \\
\qw &[5mm] \qw                  &                                                 & \qw    & \qw      & \qw      & \qw \\
\qw &[5mm] \gate{X}             & \gateinput{$a_o$} \gateoutput{$a_o \oplus y$}   & \qw    & \qw      & \meter{} & \qw \\
\end{quantikz}
\caption{Quantum algorithm for circuit diagnosis\label{fig:quantum_algorithm}}
\end{center}
\end{figure}

Figure~\ref{fig:quantum_algorithm} shows a quantum circuit that can be
used for directly computing the conditional probabilities for each
fault. Similar to many existing quantum algorithms, the idea is to put
each of the $n$ unknown fault inputs $f \in F$ in a superposition by
applying Hadamard gates. The primary inputs are initialized as
$\ket{0}$ or $\ket{1}$ corresponding to the Boolean values of the
input assignment $\alpha$.

The quantum oracle $U$ is constructed by
Algorithm~\ref{alg:make_oracle}. Before the oracle conversion the
input multi-output Boolean circuit is converted to a single-output
one. This single output circuit contains also the values of the
observed primary outputs $\beta$. Each primary outout of the
multi-output circuit is fed to the input of a multi-input
AND-gate. The outputs go directly to the inputs of the AND-gate or
through inverters, depending on the observed values for $\beta$
(primary outputs corresponding to zero bits are inverted).

As is customary for many classes of quantum algorithms, the
experiments is repeated multiple times. Each result is the collapsed
state of the system shown in Algorithm~\ref{alg:make_oracle}. It is a
probability distribution function over all possible combinations of
values for the fault inputs and the single output of the system. The
answer of the diagnostic problem is taken only from the experiments
that result in positive values for the combined output
$o$. Alternative if there is not enough probability mass in the
positive values of $o$, the result can be marginalized from
$\neg{o}$. In this case one obtains first the probability of each gate
of being healthy as opposed to being faulty.

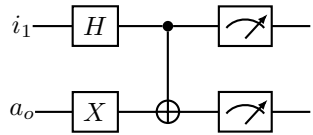
\begin{figure}[hbt]
  \centering
  \begin{quantikz}
    i_1 & \gate{H} & \ctrl{1} & \meter{} & \qw \\
    a_o & \gate{X} & \targ{}  & \meter{} & \qw \\
  \end{quantikz}
  \caption{Diagnostic quantum circuit from a single inverter oracle}
  \label{fig:quantum_inverter}
\end{figure}

Let us work out a circuit that has a single inverter with an input $i$
and an output $o$ that is known to be true. The full quantum circuit
that is used for computing $\pr{o = 1}$ is shown in
Figure~\ref{fig:quantum_inverter}.

As the circuit in Figure~\ref{fig:quantum_inverter} two qubits
only, its state can be fully described by a complex vector of size
four. Notice that in this whole paper the assumptions on the initial
state and on the unitary transformations are such that the imaginary
parts of all complex numbers are always zero.

The quantum circuit shown in Figure~\ref{fig:quantum_inverter} works
by first multiplying the unitary matrix of the quantum circuit
\[
\begin{split}
\left(X \otimes H\right)\mathrm{CNOT} &=
\frac{1}{\sqrt{2}}
\begingroup
\setlength\arraycolsep{3pt}
\begin{bmatrix}[r]
 0 &  0 &  1 &  1 \\
 0 &  0 &  1 & -1 \\
 1 &  1 &  0 &  0 \\
 1 & -1 &  0 &  0 \\
\end{bmatrix}
\begin{bmatrix}[r]
 1 &  0 &  0 &  0 \\
 0 &  0 &  0 &  1 \\
 0 &  0 &  1 &  0 \\
 0 &  1 &  0 &  0 \\
\end{bmatrix}
\endgroup
=
\\
&= \frac{1}{\sqrt{2}}
\begingroup
\setlength\arraycolsep{3pt}
\begin{bmatrix}[r]
 0 &  0 &  1 &  1 \\
 1 & -1 &  0 &  0 \\
 1 &  1 &  0 &  0 \\
 0 &  0 &  1 & -1 \\
\end{bmatrix}
\endgroup
\end{split}
\]
with the initial state
$\begingroup\setlength\arraycolsep{3pt}\begin{bmatrix}1 & 0 & 0 & 0\end{bmatrix}\endgroup^\intercal$.
This result in the final state of the quantum circuit 
$\begingroup\setlength\arraycolsep{3pt}\nicefrac{1}{\sqrt{2}}\begin{bmatrix}0 & 1 & 1 & 0\end{bmatrix}\endgroup^\intercal$.
In Dirac's notation the state is $\nicefrac{1}{\sqrt{2}}\ket{01} + \nicefrac{1}{\sqrt{2}}\ket{10}$.
After collapsing the state, the resulting histogram is $\pr{\ket{01}}
= \pr{\ket{10}} = \nicefrac{1}{2}$. Recall that the first qubit
corresponds to the switching circuit's input $i$ and the second qubit
to the value of the global output $a_o$. This means that when
computing the probability mass function of $i$ we have to take only
measurement probabilities in which $a_o = 1$. This leaves us only with
$\pr{\ket{01}} = \nicefrac{1}{2}$ from which it follows that $\pr{i =
  1} = 0$ which, given that the output is one, is consistent with the
truth table of the inverter gate.

\begin{figure}[hbt]
\centering
\begin{quantikz}
  i_1 & \gate{H} & \ctrl{2} & \qw      & \meter{} & \qw \\
  i_2 & \gate{H} & \ctrl{1} & \qw      & \meter{} & \qw \\
  a_1 & \qw      & \targ{}  & \ctrl{1} & \qw      & \qw \\
  a_o & \gate{X} & \qw      & \targ{}  & \meter{} & \qw \\
\end{quantikz}
\caption{Diagnostic quantum circuit from a single AND-gate oracle}
\label{fig:quantum_and_gate}
\end{figure}
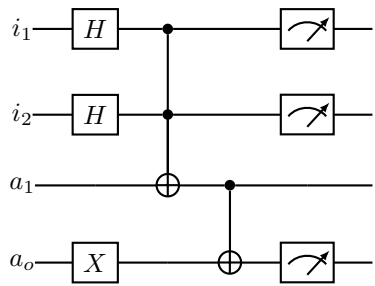

\section{Experimental Results}

Next, we perform an empirical comparison of
Algorithm~\ref{alg:sat_diag} and
Algorithm~\ref{fig:quantum_algorithm}. For the implementation of
Algorithm~\ref{alg:sat_diag} we have implemented a na\"ive model
counter based on \cadical \cite{satcompetition2020solvers}, while for
the quantum algorithm we have used \qiskit \cite{anis2021quiskit} and
its simulation engine. All experiments were performed on a $2$-CPU (4
cores per CPU) Intel Xeon \SI{3.3}{GHz} Linux computer with
\SI{1.5}{TiB} of RAM.

Several families of arithmetic circuit are diagnoses: ripple-carry
adders, ripple-borrow subtractors, barrel shifters, multipliers,
multi-operand adders, multiplexers, demultiplexers, and
comparators. The fault augmentation is stuck-at-one at gate outputs
only. For each circuit $10$ random fault injections have been
generated.

To asses the performance of the quantum approach we compare it to the
SAT-based approach, which is precise, complete, and sound. The error
in assessing the fault state of the system is defined as:
\[
\mathrm{Err} = \sum_{f \in F}{\left[\mbox{Pr}_s\left(f = 1\right) - \mbox{Pr}_q\left(f = 1\right)\right]^2},
\]
where $\mbox{Pr}_s\left(f = 1\right)$ is the probability of a fault
input $f$, computed by Algorithm~\ref{alg:sat_diag} and
$\mbox{Pr}_q\left(f = 1\right)$ is the corresponding probability
computed by Algorithm~\ref{fig:quantum_algorithm}.

\begin{figure}[hbt]
  \centering
  \begin{tikzpicture}

\begin{axis}[xlabel=Number of Gates,
             ylabel=$\mathrm{Err}$]
  \addplot+[only marks, mark=+]
  coordinates
  {
    (4, 0.001037000000000002)
    (4, 0.00023596059599999988)
    (4, 0.001660970024999999)
    (4, 0.00037700000000000065)
    (4, 0.002585557897)
    (4, 0.0001535141920000011)
    (4, 1.0040039999998945e-06)
    (4, 0.0006664201220000012)
    (4, 0.00023332562500000118)
    (4, 0.0007745645609999996)
    (8, 0.0007857064620000014)
    (8, 0.0010593293610000002)
    (8, 0.001155000000000002)
    (8, 0.001010398569999999)
    (8, 0.001873886974000002)
    (8, 0.001895664793)
    (8, 3.800000000000007e-05)
    (8, 0.0011543741180000007)
    (8, 0.0006195117260000018)
    (8, 0.00021200000000000036)
    (12, 0.0063126048460000065)
    (12, 0.05836649440700001)
    (12, 0.03194982742499998)
    (12, 0.06419745566899998)
    (12, 0.010427283023999999)
    (12, 0.05283019644499998)
    (12, 0.03725036872700002)
    (12, 0.06176916444500001)
    (12, 0.03971240250700002)
    (12, 0.03591038117400002)
    (4, 0.0007096363210000012)
    (4, 9.763416099999947e-05)
    (4, 0.0001170000000000002)
    (4, 0.00011033401600000028)
    (4, 0.00025000000000000044)
    (4, 3.071846499999965e-05)
    (4, 0.00022773828099999897)
    (4, 4.500000000000008e-05)
    (4, 0.0002770000000000005)
    (4, 0.00043366524099999846)
    (6, 4.290249999999928e-05)
    (6, 0.0016143520409999971)
    (6, 4.824369999999987e-05)
    (6, 0.0008087216799999995)
    (6, 0.00047258412100000034)
    (6, 0.0005076874760000002)
    (6, 0.0008862692190000012)
    (6, 0.0005122979560000019)
    (6, 0.0006925318560000003)
    (6, 0.0)
    (10, 0.0039142800869999975)
    (10, 0.0003906077759999981)
    (10, 0.0011469977059999981)
    (10, 0.0026162314560000013)
    (10, 0.005080900444999994)
    (10, 0.004988182473000006)
    (10, 0.0013333360010000025)
    (10, 0.00012599193799999999)
    (10, 0.003537414271999997)
    (10, 0.0008787831740000008)
    (10, 0.0007877811790000004)
    (10, 0.0018631362539999984)
    (10, 0.0022783576649999973)
    (10, 0.0019981626749999964)
    (10, 0.001600022082000001)
    (10, 0.0021195144520000024)
    (10, 0.004031669459999998)
    (10, 0.0012525369220000033)
    (10, 0.0026705046790000053)
    (10, 0.0020149066860000024)
    (20, 0.010807423393999992)
    (20, 0.006998169981000005)
    (20, 0.006782789278000006)
    (20, 0.002509665214000004)
    (20, 0.0068068277869999935)
    (20, 0.009978806440000005)
    (20, 0.0062880000000000045)
    (20, 0.004305352702999999)
    (20, 0.004547302361000001)
    (20, 0.0025493526800000016)
    (30, 0.01340287010099999)
    (30, 0.0038592952850000006)
    (30, 0.030338804065999998)
    (30, 0.013222196408999996)
    (30, 0.051661673255000005)
    (30, 0.015930567778)
    (30, 0.0033302494829999998)
    (30, 0.04430696327000001)
    (30, 0.006614406538000003)
    (30, 0.24284662125720996)
    (14, 0.0013865942789999957)
    (14, 0.0011682573939999993)
    (14, 0.005940884807999994)
    (14, 0.0020908632750000016)
    (14, 0.002659330680000002)
    (14, 0.0022640000000000043)
    (14, 0.0020260000000000035)
    (14, 0.004640603964)
    (14, 0.002387541627999995)
    (14, 0.0031705862849999876)
    (10, 0.0045438844650000036)
    (10, 0.002035593804999996)
    (10, 0.011778317724999997)
    (10, 0.0027142845320000013)
    (10, 0.0005851509540000008)
    (10, 0.002224404349999999)
    (10, 0.0005688565789999998)
    (10, 0.002448171441999999)
    (10, 0.0020656126780000013)
    (10, 0.0007735934090000014)
    (20, 0.005756531765999999)
    (20, 0.007877227472000009)
    (20, 0.023165926129000033)
    (20, 0.0045120163219999985)
    (20, 0.034012334115)
    (20, 0.004458628627999998)
    (20, 0.027615742794000007)
    (20, 0.0021555651849999953)
    (20, 0.012995650713000005)
    (20, 0.005744675135000006)
    (26, 0.01809729978000001)
    (26, 0.040804171741999996)
    (26, 1.0887347098780003)
    (26, 0.054938936480000006)
    (26, 0.06035335792800002)
    (26, 0.010998410227999994)
    (26, 0.047959100330999996)
    (26, 0.019254890486000008)
    (26, 0.10330637649400003)
    (26, 0.030322779253999985)
    (4, 0.00013312544399999982)
    (4, 0.0004801357440000019)
    (4, 0.0001841448999999992)
    (4, 0.0008650657439999997)
    (4, 0.000577000000000001)
    (4, 0.0005593617450000018)
    (4, 4.1000000000000075e-05)
    (4, 0.0007880000000000014)
    (4, 0.00010941160000000052)
    (4, 0.0004450000000000008)
    (10, 0.0023531336650000033)
    (10, 0.004011880034000003)
    (10, 0.0008917671070000026)
    (10, 0.0019140000000000034)
    (10, 0.0009422451719999984)
    (10, 0.0035834488470000017)
    (10, 0.0014862736750000016)
    (10, 0.0006010000000000011)
    (10, 0.0009348837149999994)
    (10, 0.00022361420500000253)
    (2, 3.600000000000006e-05)
    (2, 3.600000000000006e-05)
    (2, 0.0007290000000000013)
    (2, 0.0)
    (2, 0.0005290000000000009)
    (2, 0.0)
    (2, 0.00010000000000000018)
    (2, 4.9000000000000087e-05)
    (2, 3.600000000000006e-05)
    (2, 0.0)
    (4, 0.0005025218890000009)
    (4, 0.00011493984099999959)
    (4, 0.0)
    (4, 0.0004268769209999994)
    (4, 0.0001812254439999993)
    (4, 0.0017925909210000012)
    (4, 0.0)
    (4, 0.0036199475559999996)
    (4, 0.0)
    (4, 8.87363999999997e-07)
    (10, 0.0031728294930000005)
    (10, 0.0030574375630000013)
    (10, 0.0020727244899999982)
    (10, 0.005172925330000002)
    (10, 0.014221089844000008)
    (10, 0.010122304663000008)
    (10, 0.0028584237729999988)
    (10, 0.013388370247999982)
    (10, 0.0011267897010000007)
    (10, 0.0033967208389999964)
    (2, 0.00019600000000000035)
    (2, 0.0)
    (2, 0.0)
    (2, 0.0)
    (2, 0.0)
    (2, 0.0)
    (2, 0.0)
    (2, 0.0)
    (2, 0.0)
    (2, 0.0006250000000000011)
    (16, 0.015369317965000013)
    (16, 0.015432123457000003)
    (16, 0.006993370663000009)
    (16, 0.007052873706000008)
    (16, 0.002167134804000001)
    (16, 0.020207130955000017)
    (16, 0.0043700531629999985)
    (16, 0.079999982609)
    (16, 0.004330445075000004)
    (16, 0.03846169230799999)
  };
\end{axis}

\end{tikzpicture}%
  \caption{Quantum algorithm error as a function of the
    switching circuit size}
  \label{fig:system_fault_state_error}
\end{figure}
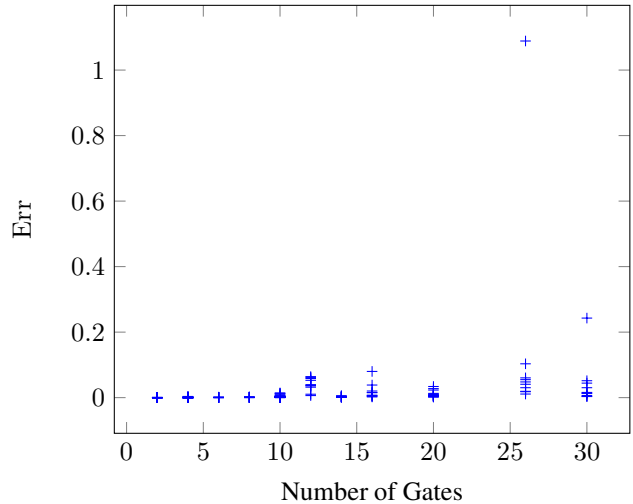

Figure~\ref{fig:system_fault_state_error} plots $\mathrm{Err}$ for a
various set of circuits from the PARC ALU benchmark. The error is
small and does not grow as a function of the circuit size. The maximum
error size is $\approx 0.08$ for a circuit with $20$ gates which
indicates that the quantum approach is suitable for approximating
circuit diagnostics.

\begin{figure}[hbt]
  \centering
  \begin{tikzpicture}

\begin{axis}[xlabel=Number of Runs (Shots) $N$,
             ylabel=$\mathrm{Err}$,
             mark options={scale=0.5}]
  \addplot+[only marks, mark=+]
  coordinates
  {
    (2, 1.2934027777777777)
    (3, 1.2517361111111112)
    (4, 0.640625)
    (5, 0.529513888888889)
    (6, 0.7184027777777778)
    (7, 0.48173611111111114)
    (8, 0.5850694444444444)
    (9, 0.2450694444444445)
    (10, 0.13715277777777776)
    (11, 0.37006944444444445)
    (12, 0.6128472222222223)
    (13, 0.25035870064279153)
    (14, 0.1344521604938272)
    (15, 0.17046889348025707)
    (16, 0.2795138888888889)
    (17, 0.13783078566732415)
    (18, 0.0920138888888889)
    (19, 0.17605583900226765)
    (20, 0.13747018313327175)
    (21, 0.19926697530864196)
    (22, 0.3519965277777778)
    (23, 0.1714891975308642)
    (24, 0.18779335949522932)
    (25, 0.06636196145124716)
    (26, 0.1474277210884354)
    (27, 0.059907856263465696)
    (28, 0.09222549245921823)
    (29, 0.1845873507805326)
    (30, 0.0952735260770975)
    (31, 0.13285337796244998)
    (32, 0.05318168934240364)
    (33, 0.036077636000840164)
    (34, 0.05079922748191977)
    (35, 0.055555555555555566)
    (36, 0.11113254458161866)
    (37, 0.13006944444444446)
    (38, 0.15640277777777778)
    (39, 0.07922322962082178)
    (40, 0.042393550557477906)
    (41, 0.07852961091293437)
    (42, 0.07336145546372819)
    (43, 0.12395833333333331)
    (44, 0.08150226251816617)
    (45, 0.10838431055709456)
    (46, 0.07197544513816626)
    (47, 0.039614898989898985)
    (48, 0.05224116161616163)
    (49, 0.10167100694444445)
    (50, 0.05889462097232366)
    (51, 0.027048405654174885)
    (52, 0.10972427679158445)
    (53, 0.0951003086419753)
    (54, 0.07630392216540864)
    (55, 0.07027654503616043)
    (56, 0.05800222255271335)
    (57, 0.07074652777777778)
    (58, 0.040252179164041176)
    (59, 0.07486536281179137)
    (60, 0.011187906952320923)
    (61, 0.05482369775257301)
    (62, 0.0924390589569161)
    (63, 0.03314948771107504)
    (64, 0.04602321370725159)
    (65, 0.04005347887454298)
    (66, 0.04693611111111112)
    (67, 0.04620277777777778)
    (68, 0.022680693999446215)
    (69, 0.04455337382425432)
    (70, 0.07872080317234284)
    (71, 0.052278517422748194)
    (72, 0.05622329059829059)
    (73, 0.034601124885215814)
    (74, 0.023925460223537153)
    (75, 0.050238715277777776)
    (76, 0.015663473376423525)
    (77, 0.026448345595506496)
    (78, 0.04250504724057584)
    (79, 0.046413845486111105)
    (80, 0.032159391534391533)
    (81, 0.03941514756944445)
    (82, 0.026359575241115075)
    (83, 0.03261804159085831)
    (84, 0.061469184027777776)
    (85, 0.047291666666666655)
    (86, 0.03174534472145276)
    (87, 0.014949972373386718)
    (88, 0.02636323563417736)
    (89, 0.03723285549254358)
    (90, 0.05025492351605724)
    (91, 0.047557902431517404)
    (92, 0.022525065746219586)
    (93, 0.04164369355177483)
    (94, 0.06207876669186194)
    (95, 0.026884040105193928)
    (96, 0.03924765312403816)
    (97, 0.01764663411048099)
    (98, 0.010486111111111104)
    (99, 0.030436847604055378)
    (100, 0.017934027777777785)
    (101, 0.02373779143037177)
    (102, 0.0341427767095729)
    (103, 0.07962057994208202)
    (104, 0.02253782022450853)
    (105, 0.02340277777777779)
    (106, 0.04224621581040498)
    (107, 0.021528498654363702)
    (108, 0.04730843240359703)
    (109, 0.03707212279847413)
    (110, 0.01066468253968254)
    (111, 0.02834045090797169)
    (112, 0.023601615140705502)
    (113, 0.020992287620557548)
    (114, 0.02201865130231977)
    (115, 0.048975694444444426)
    (116, 0.026846612559061112)
    (117, 0.024553571428571435)
    (118, 0.026350162780928398)
    (119, 0.019378793649106642)
    (120, 0.039568034689783815)
    (121, 0.009281003219153951)
    (122, 0.03276950745641673)
    (123, 0.02356804507504634)
    (124, 0.018466909360696154)
    (125, 0.031802777777777784)
    (126, 0.01638463718820861)
    (127, 0.018670578659945394)
    (128, 0.026627749372273423)
    (129, 0.040821983687288385)
    (130, 0.017796628053894502)
    (131, 0.01803611111111111)
    (132, 0.010588657691058951)
    (133, 0.03572838800911657)
    (134, 0.030799897119341578)
    (135, 0.02787023339907955)
    (136, 0.022842404509797753)
    (137, 0.023114782940081735)
    (138, 0.014081790123456788)
    (139, 0.01462407879818594)
    (140, 0.013102777777777783)
    (141, 0.01827521162137573)
    (142, 0.01157685411632262)
    (143, 0.03953742593945296)
    (144, 0.02581338908112266)
    (145, 0.03190817506695885)
    (146, 0.023245440261847956)
    (147, 0.03435904761211553)
    (148, 0.016746422542442196)
    (149, 0.01841538017223272)
    (150, 0.01924104207758054)
    (151, 0.016260370720436877)
    (152, 0.02744051051793267)
    (153, 0.040684852799521165)
    (154, 0.010979967443814324)
    (155, 0.026170287632373536)
    (156, 0.015020722817961737)
    (157, 0.020155243469936992)
    (158, 0.017643896230329713)
    (159, 0.028485172164478198)
    (160, 0.02534087685508675)
    (161, 0.01815859507165884)
    (162, 0.007073542534205835)
    (163, 0.04248050441960699)
    (164, 0.01268140173333637)
    (165, 0.02639597505668935)
    (166, 0.011426033591731253)
    (167, 0.01854962567068127)
    (168, 0.014736133878226289)
    (169, 0.01854112413194444)
    (170, 0.012472111111111107)
    (171, 0.010740055884286644)
    (172, 0.018740640701962828)
    (173, 0.011026864449483503)
    (174, 0.008356713909447989)
    (175, 0.01605493470320445)
    (176, 0.01100130544710552)
    (177, 0.022412802840434436)
    (178, 0.010940117717749722)
    (179, 0.024004247015610648)
    (180, 0.020104400255023525)
    (181, 0.02985029796232217)
    (182, 0.016538153811847915)
    (183, 0.015324634755863127)
    (184, 0.01040465205690119)
    (185, 0.013496339818321322)
    (186, 0.012388655547439325)
    (187, 0.022531721536351176)
    (188, 0.023424999999999984)
    (189, 0.014052519372961335)
    (190, 0.01645267311717716)
    (191, 0.015017918483287082)
    (192, 0.008293038408779149)
    (193, 0.018442413132514213)
    (194, 0.01506725516565542)
    (195, 0.018265036730945816)
    (196, 0.011724708102471127)
    (197, 0.005699896526182509)
    (198, 0.02042550407962749)
    (199, 0.017036716647443294)
    (200, 0.013150084182467632)
    (201, 0.013408971976590208)
    (202, 0.013523665609723873)
    (203, 0.008080418381344316)
    (204, 0.0288129364049026)
    (205, 0.012753885642399152)
    (206, 0.027570192020163194)
    (207, 0.01561091333432503)
    (208, 0.008379010219160045)
    (209, 0.03347624091100293)
    (210, 0.01210281162346478)
    (211, 0.028168402777777775)
    (212, 0.014366319444444445)
    (213, 0.010756084876104953)
    (214, 0.008728298611111122)
    (215, 0.018415798611111096)
    (216, 0.014174970952568353)
    (217, 0.00438612845415655)
    (218, 0.02817923299624199)
    (219, 0.010734964004588423)
    (220, 0.024183597337006417)
    (221, 0.0036334582852748953)
    (222, 0.012386164456637705)
    (223, 0.012018780094371496)
    (224, 0.005252767197181473)
    (225, 0.00588151927437642)
    (226, 0.022226307189542494)
    (227, 0.006363052612054645)
    (228, 0.007115645148530515)
    (229, 0.013114093689068431)
    (230, 0.005623347455080823)
    (231, 0.02186756837098693)
    (232, 0.014752401730421538)
    (233, 0.01479877222552714)
    (234, 0.025132029478458055)
    (235, 0.010441103672724885)
    (236, 0.006363552825036764)
    (237, 0.007984315650036707)
    (238, 0.009868501802522127)
    (239, 0.013557098765432096)
    (240, 0.013642851646660509)
    (241, 0.011389988604291402)
    (242, 0.008880980893343523)
    (243, 0.007686697260979079)
    (244, 0.005336111111111112)
    (245, 0.007326865225315524)
    (246, 0.01730475358294233)
    (247, 0.013134318329042993)
    (248, 0.008734795358472613)
    (249, 0.009353765276323601)
    (250, 0.006320305325987146)
    (251, 0.006999163077459787)
    (252, 0.009035794225642572)
    (253, 0.017749523066250432)
    (254, 0.008853772990677751)
    (255, 0.012249464740971761)
    (256, 0.013248953412583107)
    (257, 0.011619145957702807)
    (258, 0.0075899266107274585)
    (259, 0.008894586746263132)
    (260, 0.007785373263888888)
    (261, 0.009299408900828136)
    (262, 0.003918295062694823)
    (263, 0.010362600922722037)
    (264, 0.006441326530612247)
    (265, 0.006086696080375443)
    (266, 0.012049642464246432)
    (267, 0.013279796036416677)
    (268, 0.024219444444444453)
    (269, 0.005148260284505954)
    (270, 0.01445855034722222)
    (271, 0.011520102270594858)
    (272, 0.011633016643248483)
    (273, 0.012580636157675178)
    (274, 0.010044120129834773)
    (275, 0.00923782578875172)
    (276, 0.00870890022675736)
    (277, 0.015638775926293286)
    (278, 0.016134260382710354)
    (279, 0.004150090132775654)
    (280, 0.004579338651672494)
    (281, 0.013349206869110355)
    (282, 0.027345742781348832)
    (283, 0.010043982124469372)
    (284, 0.012208869654138484)
    (285, 0.008474805697878772)
    (286, 0.00980992887351019)
    (287, 0.00655898943622141)
    (288, 0.013685072770475974)
    (289, 0.011368317670663874)
    (290, 0.011709881747448697)
    (291, 0.013701331178824255)
    (292, 0.012555670339761248)
    (293, 0.00890644622909644)
    (294, 0.019423149149690528)
    (295, 0.009635138490657964)
    (296, 0.008621275702439065)
    (297, 0.011920439029636482)
    (298, 0.006517675154512268)
    (299, 0.005014290671341956)
  };
\end{axis}

\end{tikzpicture}%
  \caption{Quantum algorithm error as a function of the number of
    repetitions of the experiment}
  \label{fig:adder_2_sa1_001_step_shots}
\end{figure}
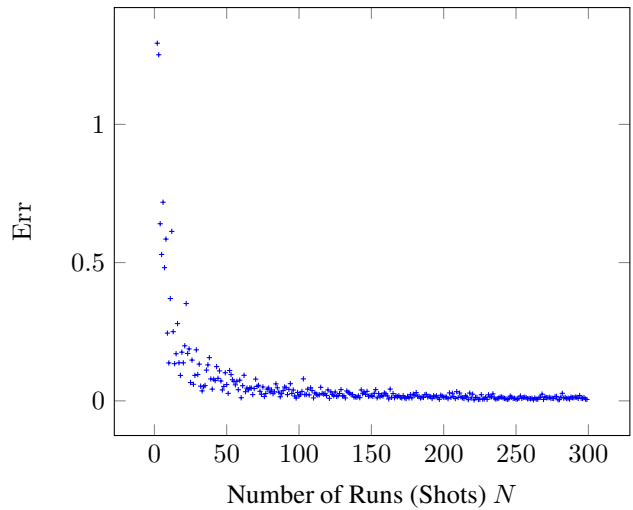

Gate-based quantum computers build the resulting probability mass
function by repeating the computation multiple times. Let us denote
the number of runs as $N$. Figure~\ref{fig:adder_2_sa1_001_step_shots}
shows the error as a function of $N$ for a two-bit full adder with
five inputs and three outputs. It is visible that the error decreases
sharply when $N$ becomes larger, and a small number of experiments is
sufficient for a good approximation.

\section{Related Work}
Our approach builds upon Deutsch–Jozsa's quantum algorithm
\cite{deutsch1992rapid} for determining if a Boolean function is
balanced or constant. The main difference between Deutsch–Jozsa's
algorithm and ours is that while Deutsch–Jozsa's algorithm is
deterministic, i.e., solves a decision problem with arbitrarily high
confidence, ours computes a probability mass function. What makes our
algorithm of real practical value is that most circuit diagnostic
problems tend to be underconstrained due to long implication
chains. While this makes classical model counting and compilation
difficult, the relatively large split between diagnoses and
inconsistent assignments results in probability mass functions with
values sufficiently different from zero or one.

The potential of quantum computing to improve the computational
performance of diagnostics has been extensively studied in the context
of quantum annealers \cite{ortiz19readiness}. An earlier and slightly
more general approach \cite{bian2016mapping} reports also on
experiments with random sampling of the solution space of a diagnostic
problem. The biggest difference between these two papers and the
algorithm presented here is that quantum annealing diagnostic
algorithms compute one diagnosis at a time while in our approach we
compute a superposition of all diagnoses.

Automated Test Pattern Generation \cite{bushnell00essentials} is
related to diagnostics of switching circuit. Singh, Bharadwaj, and
Harpreet have studied non-convential algorithms for ATPG based on DNA
and quantum computing \cite{singh2005dna}. Their approach resembles
the one presented here by the fact that they put the primary inputs of
a circuit in a super position to search for a test-vector. In their
case, however, a test-vector can be classical computed with a single
call to a SAT oracle, while in our case we need some kind of an
enumeration of multiple solutions.

\section{Discussion}
\label{sec:discussion}
On the surface, it may seem that Algorithm~\ref{fig:quantum_algorithm}
can be used for solving hard SAT or factorization problems. Indeed, we
have experimented with putting a digital multiplier circuit as an
oracle, assigning some binary number to the primary outputs and
computing the probability mass function of all inputs to the
multiplier. After running the algorithm, the probability of the single
oracle output can be used for calculating the number of satisfiable
assignments. The problem, however, is that hard SAT and factorization
problem have either one or a very small number of SAT
assignments. This non-even split of SAT/UNSAT will be essentially lost
in the error of any physical implementation of a quantum
computer. Notice that the foundational Shor \cite{shor1994algorithms}
and Deutch-Josza \cite{deutsch1992rapid} algorithms are robust to
noise, however they are closer to decision problems while we target
counting.
%
%\cite{harrow2017quantum}
%
%\cite{hogg1996quantum}
%

Quantum algorithms typically follow classical ones but that does not
have to be always the case. In this paper, for example, we have solved
large diagnostic problems, simulating a quantum computer on a
classical one. The largest circuit that we could solve, a 3-bit
subtractor, has $42$ gates and ? variables. This was challenging for
the SAT-based algorithm from Algorithm~\ref{alg:sat_diag} and it took
on average \si{?}{s} to compute all diagnoses. The good performance of
the simulated quantum algorithm is due to the fact that the overall
approach resembles compilation techniques such as Ordered Binary
Decision Diagrams (OBDDs) \cite{bryant1992symbolic}. We have also
noticed that all probabilities are rational numbers. This means that
compilation-based approach combined with Satisfiability Modulo
Theories (SMT) \cite{barrett2018satisfiability} has a lot of
potential.
\section{Conclusions}
We have presented a quantum algorithm for computing all diagnoses in a
switching circuit. The main idea is to put the bits modeling the
faults in superposition and to read out the probability of each
fault. The quantum algorithm is compared to an exact approach based on
model counting and achieves small error.

In the future we plan to improve the quantum algorithm by reducing the
number of ancillary bits in the oracle. This can be achieved by
various methods for preprocessing.
\bibliographystyle{unsrt}
\bibliography{arxiv22-qmbd}

\end{document}